\documentclass{article}
\usepackage{spconf,amsmath,graphicx}
\usepackage{multirow}
\usepackage{url}
\usepackage{color}
\usepackage{hyperref}
\ninept

\title{Image denoising using group sparsity residual and external nonlocal self-similarity prior}

\name{{Zhiyuan Zha,  Xinggan Zhang, Qiong Wang, Yechao Bai, Lan Tang}
\thanks{This work was supported by the NSFC (61571220, 61462052, 61502226) and the open research fund of National Mobile Commune. Research Lab., Southeast University (No.2015D08).}  }

\address{ {School of Electronic Science and Engineering, Nanjing University, Nanjing 210023, China.} }
\begin{document}
\maketitle
\begin{abstract}
 Nonlocal image representation has been successfully used in many image-related inverse problems including denoising, deblurring and deblocking. However, a majority of reconstruction methods only exploit the nonlocal self-similarity (NSS) prior of the degraded observation image, it is very challenging to reconstruct the latent clean image. In this paper we propose a novel model for image denoising via group sparsity residual and external NSS prior. To boost the performance of image denoising, the concept of group sparsity residual is proposed, and thus the problem of image denoising is transformed into one that reduces the group sparsity residual. Due to the fact that the groups contain a large amount of NSS information of natural images, we  obtain a good estimation of the group sparse coefficients of the original image by the external NSS prior based on Gaussian Mixture model (GMM) learning and the group sparse coefficients of noisy image is used to approximate the estimation. Experimental results have demonstrated that the proposed method not only outperforms many state-of-the-art methods, but also delivers the best qualitative denoising results with finer details and less ringing artifacts.
\end{abstract}
\section{Introduction}
\label{sec:intro}
Image denoising is not only an important problem in various image processing studies, but also an idea test bed for measuring the statistical image modeling methods. It has attracted a lot of research interest in the past few decades \cite{1,3,6,8,9,10,11,12,13,14}. Image denoising aims to estimate the latent clean image $\textbf{\emph{X}}$ from its noisy observation $\textbf{\emph{Y}}=\textbf{\emph{X}}+\textbf{\emph{V}}$, where $\textbf{\emph{V}}$ is usually assumed to be additive white Gaussian noise. Due to the ill-posed nature of image denoising, it is critical to exploit the prior knowledge that characterizes the statistical features of the images.

Previous models mainly employed priors on level of pixel, such as  Tikhonov regularization \cite{2}, total variation (TV) regularization \cite{3}.  These methods are effective in removing the noise artifacts but smear out  details and tend to oversmooth the images \cite{4,5}.

Recently, patch-based prior have shown promising performance in image denoising. One representative example is sparse coding based scheme, which assumes that each patch of an image can be precisely represented by a sparse coefficient vector whose entries are mostly zero or close to zero \cite{6,11,5,7}. Considering that natural images are non-Gaussian and image patches are regarded as samples of a multivariate vector, Gaussian mixture models (GMMs) have emerged as favored prior for natural image patches in various image restoration studies \cite{8,9,10,11}. However, patch-based model usually suffers from some limits, such as great computational complexity, neglecting the relationship among similar patches \cite{16,24}.

Inspired by the fact that natural images contain a large number of mutually similar patches at different locations, this so-called nonlocal self-similarity (NSS) prior was initially utilized in the work of nonlocal mean denoising\cite{1}. Due to its effectiveness, a large amount of further developments \cite{11,12,13,14,4,5,7,16,24,15} have been proposed. For instance, a very popular method is BM3D \cite{12}, which exploited nonlocal similar 2D image patches and 3D transform domain collaborative filtering.  In \cite{13}, Marial $\emph{et al}.$ advanced the idea of NSS by group sparse coding.  LPG-PCA \cite{14} uses nonlocal similar patches as data samples to estimate statistical parameters for PCA training. In practical, these methods belong to the category of group sparsity.

Though group sparsity has verified its great success in image denoising, all of above methods only exploited the NSS prior of noisy input image.  However, only considering the NSS prior of noisy input image, it is very challenging to recover the latent clean image from noisy observation.

With the above considerations, in this work we propose a novel method for image denoising with group sparsity residual and external NSS prior. The contribution of this paper is as follows.  First, to improve the performance of image denoising, we propose the concept of group sparsity residual, and thus the problem of image denoising turned into reducing the group sparsity residual. Second, due to the fact that the groups contain a large amount of NSS information of natural images, to reduce  residual, we  obtain a good estimation of the group sparse coefficients of the original image by the NSS prior of natural images based on Gaussian Mixture model (GMM) learning and the group sparse coefficients of noisy input image is used to approximate the estimation.  Our experimental results have demonstrated that the proposed method outperforms many state-of-the-art methods. What's more, the proposed method delivers the best qualitative denoising results with finer details and less ringing artifacts.

\section{Background}
\label{sec:format}
\subsection {Group-based sparse coding\label{sub:2.1.}}
Recent studies \cite{12,13,16,15} have revealed that structured or group sparsity can offer powerful reconstruction performance for image denoising. To be concrete, given a clean image $\textbf{\emph{X}}$, for each image patch $\textbf{\emph{x}}$ of size $d\times d$ in $\textbf{\emph{X}}$, its $m$ best matched  patches are selected from a $W \times W$ sized search window to form a group ${\textbf{\emph{X}}}_i$, denoted by ${\textbf{\emph{X}}}_i=\{{\textbf{\emph{x}}}_{i,1}, {\textbf{\emph{x}}}_{i,2},...,{\textbf{\emph{x}}}_{i,m}\}$, where ${\textbf{\emph{x}}}_{i,m}$ denotes the $m$-th similar patch (column vector) of the $i$-th group. Similar to patch-based sparse coding \cite{6,5,7}, given a dictionary ${\textbf{\emph{D}}}_{i}$, each group ${\textbf{\emph{X}}}_i$ can be sparsely represented as ${\textbf{\emph{B}}}_i={\textbf{\emph{D}}}_i^T{\textbf{\emph{X}}}_i$ and solved by the following $\ell_p$-norm minimization problem,
\begin{equation}
{{\textbf{\emph{B}}}_i}=\arg\min\nolimits_{{\textbf{\emph{B}}}_i} \{||{\textbf{\emph{X}}}_i-{\textbf{\emph{D}}}_i{{\textbf{\emph{B}}}_i}||_F^2+\lambda||{{\textbf{\emph{B}}}_i}||_p\}
\label{eq:1}
\end{equation} 
where $\lambda$ is the regularization parameter, and $p$ characterizes the sparsity of ${{\textbf{\emph{B}}}_i}$. Then the whole image ${\textbf{\emph{X}}}$ can be represented by the set of group sparse codes $\{{{\textbf{\emph{B}}}_i}\}$.

In image denoising,  each noise patch $\textbf{\emph{y}}$ is extracted from the  noisy image ${\textbf{\emph{Y}}}$, like in the above paragraph, we search for its similar patches to generate a group ${\textbf{\emph{Y}}}_i$, i.e., ${\textbf{\emph{Y}}}_i=\{{\textbf{\emph{y}}}_{i,1}, {\textbf{\emph{y}}}_{i,2},...,{\textbf{\emph{y}}}_{i,m}\}$. Thus, image denoising then is turned to how to recover ${\textbf{\emph{X}}}_i$ from ${\textbf{\emph{Y}}}_i$ by using group sparse coding,
\begin{equation}
{{\textbf{\emph{A}}}_i}=\arg\min\nolimits_{{\textbf{\emph{A}}}_i} \{||{\textbf{\emph{Y}}}_i-{\textbf{\emph{D}}}_i{{\textbf{\emph{A}}}_i}||_F^2+\lambda||{{\textbf{\emph{A}}}_i}||_p\}
\label{eq:2}
\end{equation} 

Once all group sparse codes $\{{{\textbf{\emph{A}}}_i}\}$ are achieved, the latent clean image $\textbf{\emph{X}}$ can be reconstructed as $\hat{\textbf{\emph{X}}}={\textbf{\emph{D}}}{\textbf{\emph{A}}}$.
\section{Image denoising using group sparsity residual with external NSS prior}
\label{sec:pagestyle}
As we know, only considering the NSS prior of noisy input image, it is very challenging to recover the latent clean image from noisy observation image. In this section, to improve the performance of image denoising, we propose the concept of group sparsity residual, and  thus the problem of image denoising is translated into reducing the group sparsity residual. Due to the fact that groups possess a large amount of NSS information of natural images, to reduce residual, a good estimation of the group sparse coefficients of the original image is obtained by the external NSS prior based on GMM learning and the group sparse coefficients of noisy image is used to approximate the estimation.
\subsection {Group sparsity residual\label{sub:3.1.}}
Although group sparsity has demonstrated its effectiveness in image denoising, due to the influence of noise, it is very difficult to estimate the true group sparse codes $\textbf{\emph{B}}$ from noisy image $\textbf{\emph{Y}}$. In other words, the group sparse codes ${{\textbf{\emph{A}}}}$ obtained by solving Eq.~\eqref{eq:2} are expected to be close enough to the true group sparse codes ${{\textbf{\emph{B}}}}$ of the original image ${{\textbf{\emph{X}}}}$. As a consequence, the quality of image denoising largely depends on the level of the group sparsity residual,  which is defined as the difference between group sparse codes ${{\textbf{\emph{A}}}}$ and true group sparse codes ${{\textbf{\emph{B}}}}$,
\begin{equation}
{{\textbf{\emph{R}}}}={{\textbf{\emph{A}}}}-{{\textbf{\emph{B}}}}
\label{eq:3}
\end{equation} 

Therefore, to reduce the group sparsity residual ${{\textbf{\emph{R}}}}$ and enhance the accuracy of  ${{\textbf{\emph{A}}}}$, we propose the group sparsity residual  to image denoising, Eq.~\eqref{eq:2} can be rewritten as
\begin{equation}
{{\textbf{\emph{A}}}_i}=\arg\min\nolimits_{{\textbf{\emph{A}}}_i} \{||{\textbf{\emph{Y}}}_i-{\textbf{\emph{D}}}_i{{\textbf{\emph{A}}}_i}||_F^2+\lambda||{{\textbf{\emph{A}}}_i}-{{\textbf{\emph{B}}}_i}||_p\}
\label{eq:4}
\end{equation} 

However, it can be seen that the true group sparse codes ${{\textbf{\emph{B}}}}$ and $p$ are unknown because the original image ${{\textbf{\emph{X}}}}$ is not available. We now discuss how to obtain ${{\textbf{\emph{B}}}}$ and $p$.
\subsection {How to estimate the true group sparse codes ${\textbf{\emph{B}}}$}
Since the original image ${{\textbf{\emph{X}}}}$ is not available, it seems to be difficult to obtain the true group sparse codes ${{\textbf{\emph{B}}}}$. Nonetheless, we can compute some good estimation of ${\textbf{\emph{B}}}$. Generally speaking, there are various methods to estimate the true group sparse codes ${{\textbf{\emph{B}}}}$, which depends on the prior knowledge of ${{\textbf{\emph{B}}}}$ we have. In recent years, patch-based or group-based priors referring to denoising operators learned from natural images achieved the state-of-the-art denoising results \cite{6,8,11}. For instance, in \cite{6}, a dictionary learning-based method is introduced for compact patch representation, whereas in \cite{11}, a GMM model is learned from natural image groups  based on NSS scheme and used as a prior for denoising.  Due to the fact that the  groups contain a rich amount of NSS information of natural images, we can achieve a good estimation of ${\textbf{\emph{B}}}$ by the NSS prior of natural images based on GMM learning.
\subsubsection{Learning the NSS prior from natural images by GMM}
Like in subsection~\ref{sub:2.1.}, we extract $n$ groups from a given clean natural image dataset, and we denote one group as
\begin{equation}
{\overline{\textbf{\emph{Z}}}}_{\emph{i}}={\{{\bar{\textbf{\emph{z}}}}_{\emph{i, j}}\}}_{\emph{j}=1}^{\emph{m}}, \ \emph{i}=1, 2, ..., \emph{n}
\label{eq:5}
\end{equation} 
where  $ {\overline{\textbf{\emph{Z}}}}_{\emph{i}}$ \footnote{The advantage of group mean substraction is that it can further promote the NSS prior learning because the possible number of patterns is reduced, while the training samples of each pattern are increased.}is the group mean substraction of each group ${\textbf{\emph{Z}}}_i$ and ${\overline{\textbf{\emph{z}}}}_{i,j}$ denotes the $j$-th similar patch (column vector) of the $i$-th group. Since GMM has been successfully used to model the image patch or group priors such as EPLL \cite{8}, PLE \cite{9} and PGPD \cite{11}, we adopt the strategy in \cite{11} and learn a finite GMM over natural image groups $ {\{{\overline{\textbf{\emph{Z}}}}_{\emph{i}}\}}$  as group priors. By using the GMM model, the likelihood of a given group $ {\{{\overline{\textbf{\emph{Z}}}}_{\emph{i}}\}}$  is:
\begin{equation}
\begin{aligned}
P({\overline{\textbf{\emph{Z}}}}_{\emph{i}})=\sum\nolimits_{k=1}^K \pi_k\prod\nolimits_{j=1}^m {\mathcal{N}}({\bar{\textbf{\emph{z}}}}_{\emph{i, j}}|\boldsymbol\mu_k, \boldsymbol\Sigma_k)
\label{eq:6}
\end{aligned}
\end{equation} 
where $K$ is the total number of mixture components selected from the GMM, and the GMM model is parameterized by mean vectors $\{{\boldsymbol\mu}_k\}$, covariance matrices $\{{\boldsymbol\Sigma}_k\}$, and mixture weights of mixture components $\{{\pi}_k\}$. By assuming that all the groups are independent, the overall objective likelihood function is ${\mathcal{L}}=\Pi_{i=1}^n P({\overline{\textbf{\emph{Z}}}}_{\emph{i}})$. Then by applying log to it, we maximize the  objective function as:
\begin{equation}
\begin{aligned}
{\rm ln} \ {\mathcal{L}}= \sum\nolimits_{i=1}^n {\rm ln}(\sum\nolimits_{k=1}^K \pi_k\prod\nolimits_{j=1}^m {\mathcal{N}}({\bar{\textbf{\emph{z}}}}_{\emph{i, j}}|\boldsymbol\mu_k, \boldsymbol\Sigma_k)
\label{eq:7}
\end{aligned}
\end{equation} 
We collectively represent three parameters ${\boldsymbol\mu}_k, {\boldsymbol\Sigma}_k$ and ${\pi}_k$ by $\boldsymbol\Theta=\{{\boldsymbol\mu}_k, {\boldsymbol\Sigma}_k, {\pi}_k\}_{k=1}^K$, and $\boldsymbol\Theta$ is learned using Expectation Maximization algorithm (EM) \cite{8,11,17}. For more details about EM algorithm, please refer to \cite{17}.

Thus, for each noisy group ${{\textbf{\emph{Y}}}}_i$ \footnote{All noisy groups are preprocessed by mean substraction. The mean $\boldsymbol\mu_i$ of each noisy group ${{\textbf{\emph{Y}}}}_i$ is very close to the mean of the original group ${{\textbf{\emph{X}}}}_i$ because the mean of noise ${\textbf{\emph{V}}}$ is nearly zero. Thus, the mean $\boldsymbol\mu_i$ can be added back to the denoised group  $\hat{{{\textbf{\emph{X}}}}}_i$ to achieve the latent clean image $\hat{{\textbf{\emph{X}}}}$.} of noisy input image ${{\textbf{\emph{Y}}}}$, the best suitable Gaussian component is selected from this group-based GMM learning stage. Specifically, assume that the image is corrupted by the  Gaussain white noise with variance $\sigma^2$, then the covariance matrix of the $k$-th Gaussian component will turn into $\boldsymbol\Sigma_k+\sigma^2 \textbf{\emph{I}}$, where $\textbf{\emph{I}}$ represents the identity matrix. The selection that ${{\textbf{\emph{Y}}}}_i$ belongs to the $k$-th Gaussian component can be accomplished by computing the following posterior probability,
\begin{equation}
\begin{aligned}
P(k|{{\textbf{\emph{Y}}}}_i)= \frac{\prod_{j=1}^m {\mathcal{N}}({{\textbf{\emph{y}}}}_{\emph{i, j}}|\textbf{0}, \boldsymbol\Sigma_k+\sigma^2 \textbf{\emph{I}})}{\sum_{l=1}^K {\prod_{j=1}^m {\mathcal{N}}({{\textbf{\emph{y}}}}_{\emph{i, j}}|\textbf{0}, \boldsymbol\Sigma_l+\sigma^2 \textbf{\emph{I}})}}
\label{eq:8}
\end{aligned}
\end{equation} 

We maximize it, and finally, the Gaussian component with the highest probability is selected to operate each group ${{\textbf{\emph{Y}}}}_i$.

Then, we assume that the $k$-th Gaussian component is selected for the group ${{\textbf{\emph{Y}}}}_i$. Actually, GMM model is equivalent to the block sparse estimation with a block dictionary having $K$ blocks wherein each block corresponds to the PCA basis of one of the Gaussian components in the mixture \cite{9,18}. Thus,  the covariance matrix of the $k$-th Gaussian component is denoted by $\boldsymbol\Sigma_k$. By  using singular value decomposition to $\boldsymbol\Sigma_k$, we have
\begin{equation}
\boldsymbol\Sigma_k={\textbf{\emph{U}}}_k \boldsymbol\Lambda_k {{\textbf{\emph{U}}}_k}^T
\label{eq:9}
\end{equation} 
where ${\textbf{\emph{U}}}_k$ is an orthonormal matrix formed by the eigenvector of $\boldsymbol\Sigma_k$ and $\boldsymbol\Lambda_k$ is the diagonal matrix of eigenvalues. With the group-based GMM learning, as the statistical structures of NSS variations in natural image are captured by the eigenvectors in ${\textbf{\emph{U}}}_k$, and thus ${\textbf{\emph{U}}}_k$ can be used to represent the structural variations of the groups in that component. Finally, for each group ${{\textbf{\emph{Y}}}}_i$, the  true group sparse code ${{\textbf{\emph{B}}}}_i$ can be estimated by ${{\textbf{\emph{B}}}}_i= {{\textbf{\emph{U}}}_k}^{-1}{{\textbf{\emph{Y}}}}_i$.

Similar to ${{\textbf{\emph{B}}}}_i$,  the covariance matrix of each group ${{\textbf{\emph{Y}}}}_i$ is defined as $\boldsymbol\Sigma_i$ and we have
\begin{equation}
\boldsymbol\Sigma_i={\textbf{\emph{D}}}_i \boldsymbol\Lambda_i {{\textbf{\emph{D}}}_i}^T
\label{eq:10}
\end{equation} 
where ${\textbf{\emph{D}}}_i$ is an orthonormal matrix formed by the eigenvector of $\boldsymbol\Sigma_i$ and $\boldsymbol\Lambda_i$ is the diagonal matrix of eigenvalues.  Thus, ${{\textbf{\emph{A}}}}_i$ can be solved by ${{\textbf{\emph{A}}}}_i= {{\textbf{\emph{D}}}_i}^{-1}{{\textbf{\emph{Y}}}}_i$.
\begin{figure}[t]
\begin{minipage}[b]{1\linewidth}
  \centering
  \centerline{\includegraphics[width=8cm]{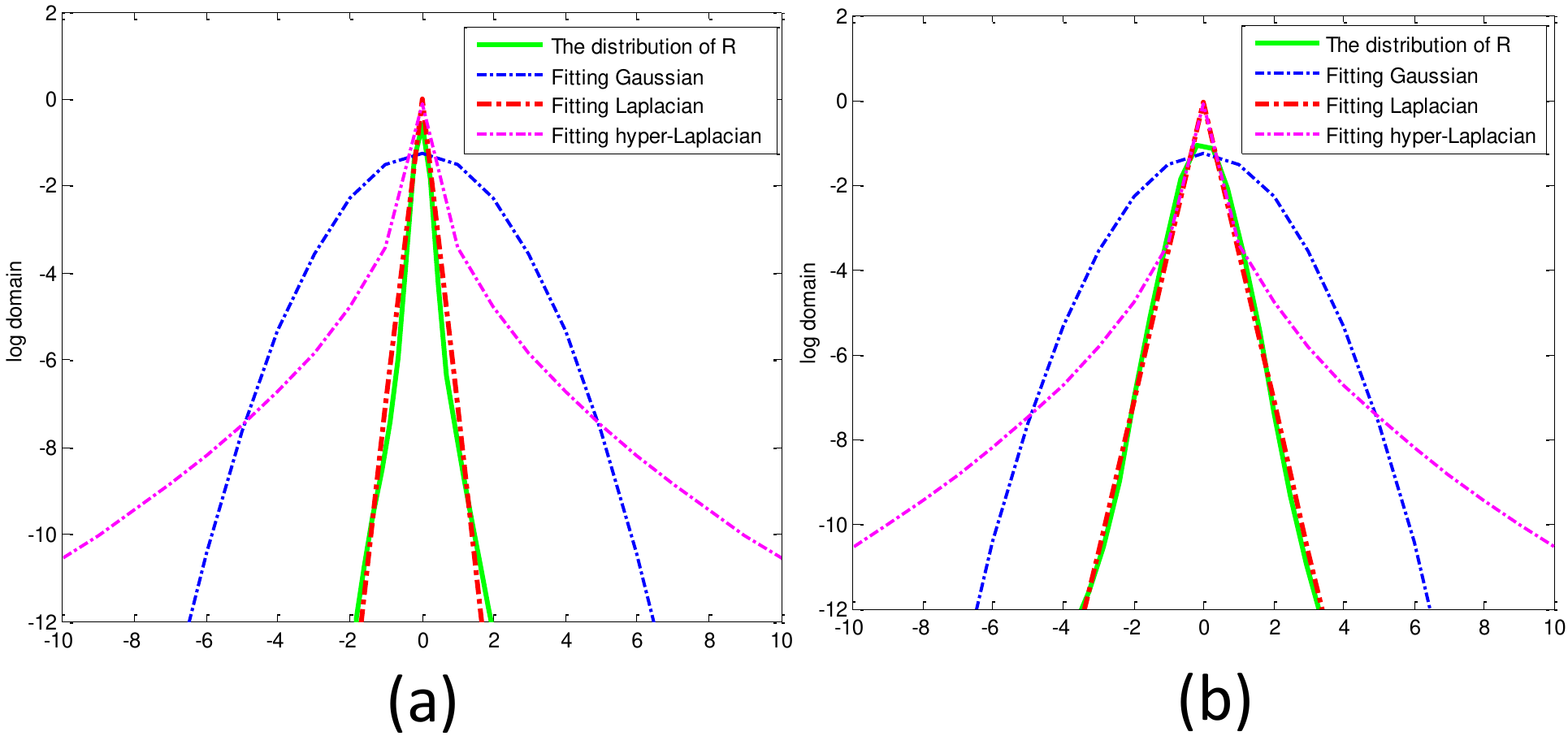}}
\end{minipage}
\caption{The distribution of $\textbf{\emph{R}}$, fitting Gaussian, Laplacian and hyper-Laplacian distribution for image $\emph{Monarch}$ with $\sigma=30$ in (a) and $\emph{foreman}$ with $\sigma=100$ in (b).}
\label{fig:1}
\end{figure}
\subsection {How to determine $p$}
Besides estimating ${{\textbf{\emph{B}}}}$, we also need to determine the value of $p$. Here we perform some experiments to investigate the statistical property of ${{\textbf{\emph{R}}}}$, where ${{\textbf{\emph{R}}}}$ denotes the set of ${{\textbf{\emph{R}}}}_i={{\textbf{\emph{A}}}}_i-{{\textbf{\emph{B}}}}_i$. In these experiments,  image $\emph{Monarch}$ and $\emph{foreman}$ are used as  examples, where $\emph{Monarch}$ and $\emph{foreman}$ are added by Gaussian white noise with standard deviation $\sigma$= 30 and $\sigma$= 100, respectively. We plot the histogram of ${{\textbf{\emph{R}}}}$ as well as the fitting Gaussian, Laplacian and hyper-Laplacian distribution in the log domain in Fig.~\ref{fig:1}(a) and Fig.~\ref{fig:1}(b), respectively. It can be seen that the histogram of ${{\textbf{\emph{R}}}}$ can be well characterized  by the Laplacian distribution. Thus, the $\ell_1$-norm is adopted to regularize ${{\textbf{\emph{R}}}_i}$, and Eq.~\eqref{eq:4} can be rewritten as
\begin{equation}
\begin{aligned}
{{\textbf{\emph{A}}}_i}& = \arg\min\nolimits_{{\textbf{\emph{A}}}_i} \{||{\textbf{\emph{Y}}}_i-{\textbf{\emph{D}}}_i{{\textbf{\emph{A}}}_i}||_F^2+\lambda||{{\textbf{\emph{A}}}_i}-{{\textbf{\emph{B}}}_i}||_1\}\\
& =\arg\min\nolimits_{\tilde{\boldsymbol\alpha}_i} \{||\tilde{{\textbf{\emph{y}}}}_i-\tilde{{\textbf{\emph{D}}}}_i{\tilde{\boldsymbol\alpha}_i}||_2^2+\lambda||{\tilde{\boldsymbol\alpha}_i}-{\tilde{\boldsymbol\beta}_i}||_1\}
\label{eq:11}
\end{aligned}
\end{equation} 
where $\tilde{{\textbf{\emph{y}}}}_i, {\tilde{\boldsymbol\alpha}_i}$, and ${\tilde{\boldsymbol\beta}_i}$ denote the vectorization of the matrix ${{\textbf{\emph{Y}}}_i}, {{\textbf{\emph{A}}}_i}$ and ${{\textbf{\emph{B}}}_i}$, respectively. Each column $\tilde{{\textbf{\emph{d}}}}_h$ of the matrix $\tilde{{\textbf{\emph{D}}}}_i=[\tilde{{\textbf{\emph{d}}}}_1, \tilde{{\textbf{\emph{d}}}}_2, ..., \tilde{{\textbf{\emph{d}}}}_J]$  denotes the vectorization of the rank-one matrix.
\subsection{How to solve Eq.~\eqref{eq:11} }
For fixed ${\tilde{\boldsymbol\beta}_i}$ and $\lambda$, it can be seen that Eq.~\eqref{eq:11} is convex and can be solved efficiently by using some iterative thresholding algorithms. We adopt the surrogate algorithm in \cite{19} to solve Eq.~\eqref{eq:11}. In the $t+1$-iteration, the proposed shrinkage operator can be calculated as
\begin{equation}
{\tilde{\boldsymbol\alpha}_i}^{t+1}= {{{\textbf{\emph{S}}}}_{\lambda}}({\tilde{{\textbf{\emph{D}}}}_i}^{-1}{{\hat{\tilde{{\textbf{\emph{x}}}}}}_i}^{t}-{{\tilde{\boldsymbol\beta}_i}}^{t})+{{\tilde{\boldsymbol\beta}_i}}^{t}
\label{eq:12}
\end{equation} 
where ${{{\textbf{\emph{S}}}}_{\lambda}}(\cdot)$ is the soft-thresholding operator, ${{\hat{\tilde{{\textbf{\emph{x}}}}}}_i}$  represents the vectorization of the $i$-th reconstructed group ${{\hat{{\textbf{\emph{X}}}}}_i}$. The above shrinkage operator follows the standard surrogate algorithm, and more details can be seen in \cite{19}.

The parameter $\lambda$ that balances the fidelity term and the regularization term should be adaptively determined for better denoising performance. Inspired by \cite{20}, the regularization parameter $\lambda_i$ of each group ${\textbf{\emph{Y}}}_i$ is set as ${{{\lambda}_{i}}}={c*2\sqrt{2}{\sigma}^2}/{{{\sigma}_{i}}}$, where ${{\sigma}_{i}}$ denotes the estimated variance of ${\textbf{\emph{R}}}_i$, and $c$ is a small constant.

After obtaining the  solution ${{\textbf{\emph{A}}}_i}$ in Eq.~\eqref{eq:12}, the clean group ${{\textbf{\emph{X}}}_i}$ can be reconstructed as ${{\hat{\textbf{\emph{X}}}}_i}={{\textbf{\emph{D}}}_i}{{\textbf{\emph{A}}}_i}$. Then the clean image ${{\hat{\textbf{\emph{X}}}}}$ can be reconstructed by aggregating all the group ${{\hat{\textbf{\emph{X}}}}_i}$.

Besides, we could execute the above denoising procedures for better results after several iterations. In the $t$+1-th iteration, the iterative regularization strategy \cite{21} is used to update the estimation of noise variance. Then the standard divation of noise in $t$+1-th iteration is adjusted as ${(\sigma^{t+1})}=\gamma*\sqrt{({\sigma^2-||{{\textbf{\emph{Y}}}}-{\hat{{\textbf{\emph{X}}}}}^{t+1}||_2^2})}$, where $\gamma$ is a constant.

According to the above analysis, it can be seen that the proposed model employs the group sparsity residual and external NSS prior for image denoising. The proposed denoising procedure is summarized in $\textbf{Algorithm 1}$.
\begin{table}[t]
\centering  
\begin{tabular}{lccc}  
\hline  
\qquad \qquad   $\textbf{Algorithm 1}$: The proposed denoising algorithm\\
\hline
$\textbf{Input:}$ \ Noisy image ${{\textbf{\emph{Y}}}}$ and  Group-based GMM learning model.\\

  $\rm \textbf{Initialization:} \  {\hat{{\textbf{\emph{X}}}}}={{\textbf{\emph{Y}}}}, \emph{c}, \emph{m}, \emph{d}, \emph{W}, \emph{K}, \sigma, \gamma, \rho$;\\
  $\rm \textbf{For}$\ $t=1, 2, ..., Iter$ $\rm \textbf{do}$\\
   \ \ \ Iterative regularization ${{\textbf{\emph{Y}}}}^{t+1}= {\hat{{\textbf{\emph{X}}}}}^{t}+\rho({{\textbf{\emph{Y}}}}-{\hat{{\textbf{\emph{X}}}}}^{t})$;\\
 \ \ \ \ \ $\rm \textbf{For}$\ each patch ${{\textbf{\emph{y}}}}$ in ${{\textbf{\emph{Y}}}}$ $\rm \textbf{do}$\\
 \ \ \ \ \ \ \ \ Find a group ${{{\textbf{\emph{Y}}}}_i}^{t+1}$ for each patch ${{\textbf{\emph{y}}}}$.\\
\ \ \ \ \ \ \ \ The best Gaussian component is selected by Eq.~\eqref{eq:8}.\\
\ \ \ \ \ \ \ \ Constructing dictionary ${{{\textbf{\emph{U}}}}_k}$ by Eq.~\eqref{eq:9}.\\
\ \ \ \ \ \ \ \ Update ${{{\textbf{\emph{B}}}}_i}^{t+1}$ computing by ${{\textbf{\emph{B}}}}_i={{{\textbf{\emph{U}}}}_k}^{-1}{{\textbf{\emph{Y}}}}_i$.\\
\ \ \ \ \ \ \ \ Constructing dictionary ${{{\textbf{\emph{D}}}}_i}^{t+1}$ by Eq.~\eqref{eq:10}.\\
\ \ \ \ \ \ \ \ Update ${{{\textbf{\emph{A}}}}_i}^{t+1}$ computing by ${{\textbf{\emph{A}}}}_i={{{\textbf{\emph{D}}}}_i}^{-1}{{\textbf{\emph{Y}}}}_i$.\\
\ \ \ \ \ \ \ \ Update ${\lambda_i}^{t+1}$ computing by ${{{\lambda}_{i}}}={c*2\sqrt{2}{\sigma}^2}/{{{\sigma}_{i}}}$.\\
\ \ \ \ \ \ \ \ Update ${{{\textbf{\emph{A}}}}_i}^{t+1}$ computing by Eq.~\eqref{eq:12}.\\
\ \ \ \ \ \ \ \ Get the estimation ${{{\textbf{\emph{X}}}}_i}^{t+1}$ =${{{\textbf{\emph{D}}}}_i}^{t+1}$${{{\textbf{\emph{A}}}}_i}^{t+1}$.\\
\ \ \ \ \ $\rm \textbf{End for}$\\
\ \ \ Aggregate ${{{\textbf{\emph{X}}}}_i}^{t+1}$ to form the recovered image ${\hat{{\textbf{\emph{X}}}}}^{t+1}$.\\
  $\rm \textbf{End for}$\\
     $\textbf{Output:}$ ${\hat{\textbf{\emph{X}}}}^{t+1}$.\\
\hline
\end{tabular}
\end{table}

\section {Experimental Results\label{sec:4}}
\begin{figure}[!htbp]
\begin{minipage}[b]{1\linewidth}
  \centering
  \centerline{\includegraphics[width=8.5cm]{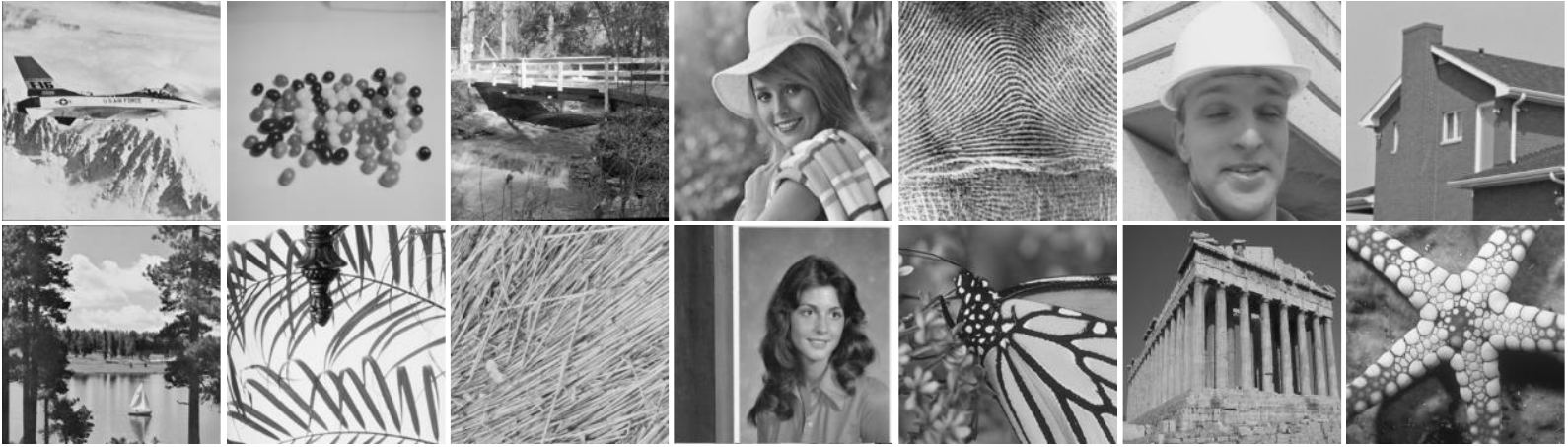}}
\end{minipage}
\caption{All test images.}
\label{fig:2}
\end{figure}
In this section, we validate the performance of the proposed denoising algorithm and compare it with several state-of-the-art denoising methods, including BM3D \cite{12}, EPLL \cite{8}, NCSR \cite{7}, GID \cite{22}, LINE \cite{10}, PGPD \cite{11} and aGMM \cite{23}. We evaluate the competing methods on 14 typical natural images, whose scene are displayed in Fig.~\ref{fig:2}. The training groups used in our experiments were sampled based on NSS scheme from the Kodak photoCD dataset\footnote{http://r0k.us/graphics/kodak/.}. The detailed setting of parameters are shown in Table~\ref{tab:1}, including the number of Gaussian components $K$ \footnote{ Since possible patterns and variables to learn can be reduced, the number of Gaussian component is not necessarily large.}, search window $W$, the number of similar patches in a group $m$, patch size $d$ and $c$, $\gamma$, $\rho$. 
\begin{table}[!htbp]
\caption{Parameter settings}
\scriptsize
\centering  
\begin{tabular}{|c|c|c|c|c|c|c|c|c|}
\hline
\multicolumn{1}{|c|}{}&\multicolumn{4}{|c|}{GMM Learning Stage}&\multicolumn{3}{|c|}{Denoising Stage}\\
\hline
\multirow{1}{*}{{{Noise level}}}&{\emph{{K}}}&{\emph{{W}}}&{\emph{{d}}}
&{\emph{{m}}}&{\emph{{c}}}&${\rho}$&${\gamma}$\\
 \hline
 \multirow{1}{*}{$\sigma\leq10$}        & 64 &  50  & 6 & 80  & 0.14 & 0.19  & 1.08\\
 \hline
 \multirow{1}{*}{$10<\sigma\leq 20$}    & 64 &  50  & 6 & 80  & 0.13 & 0.20  & 1.05\\
\hline
 \multirow{1}{*}{$20<\sigma\leq 30$}    & 64 &  50  & 7 & 90  & 0.12 & 0.21  & 1.05\\
\hline
 \multirow{1}{*}{$30<\sigma\leq 40$}    & 64 &  50  & 8 & 100 & 0.11 & 0.22  & 1.05\\
\hline
 \multirow{1}{*}{$40<\sigma\leq 50$}    & 64 &  50  & 8 & 100  & 0.10 & 0.23  & 1.05\\
\hline
 \multirow{1}{*}{$50<\sigma\leq 75$}    & 64 &  50  & 9 & 120 & 0.09 & 0.24 & 1.00\\
\hline
 \multirow{1}{*}{$75<\sigma\leq 100$}   & 64 &  50  & 9 & 120  & 0.08 & 0.25  & 1.00\\
\hline
\end{tabular}
\label{tab:1}
\end{table}

\begin{figure}[!htb]
\begin{minipage}[b]{1\linewidth}
  \centering
  \centerline{\includegraphics[width=9cm]{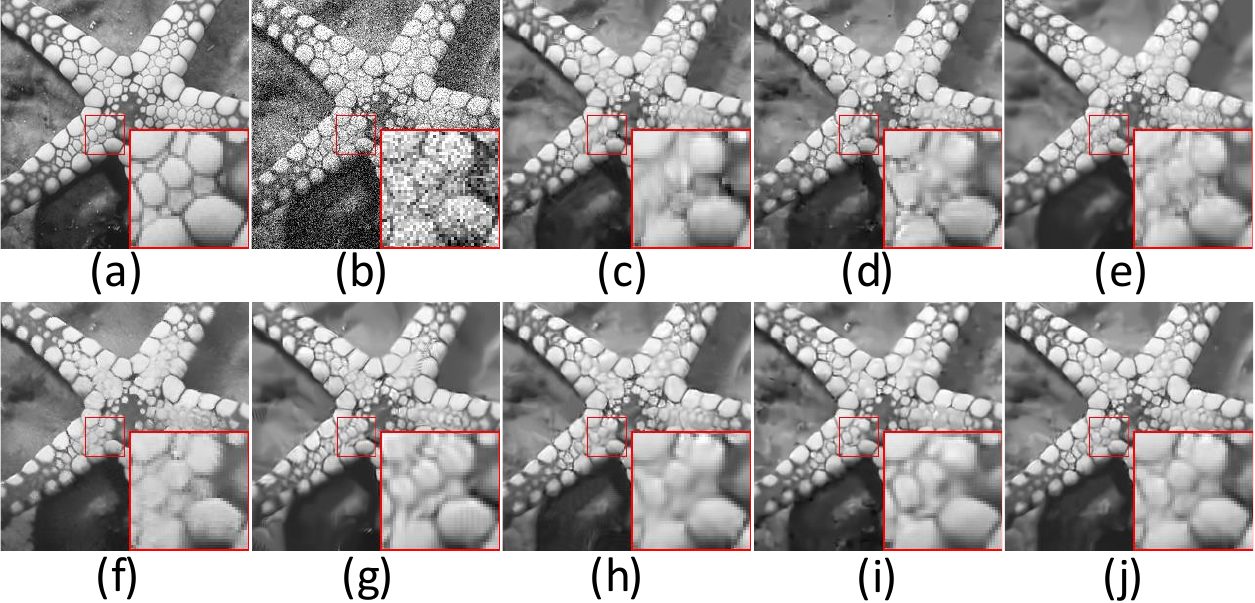}}
\end{minipage}
\caption{Denoising results on image $\emph{starfish}$ by different methods (noise level $\sigma=40$). (a) Original image; (b) Noisy image;  (c) BM3D \cite{12} (PSNR=25.94dB); (d) EPLL \cite{8} (PSNR=26.07dB); (e) NCSR \cite{7} (PSNR=26.06dB); (f) GID \cite{22} (PSNR=25.39dB); (g) LINE  \cite{10}(PSNR=26.19dB); (h) PGPD \cite{11} (PSNR=26.21dB); (i) aGMM \cite{23} (PSNR= 26.25dB); (j) Proposed (PSNR=\textbf{26.57dB}).}
\label{fig:3}
\end{figure}

\begin{figure}[!htb]
\begin{minipage}[b]{1\linewidth}
  \centering
  \centerline{\includegraphics[width=9cm]{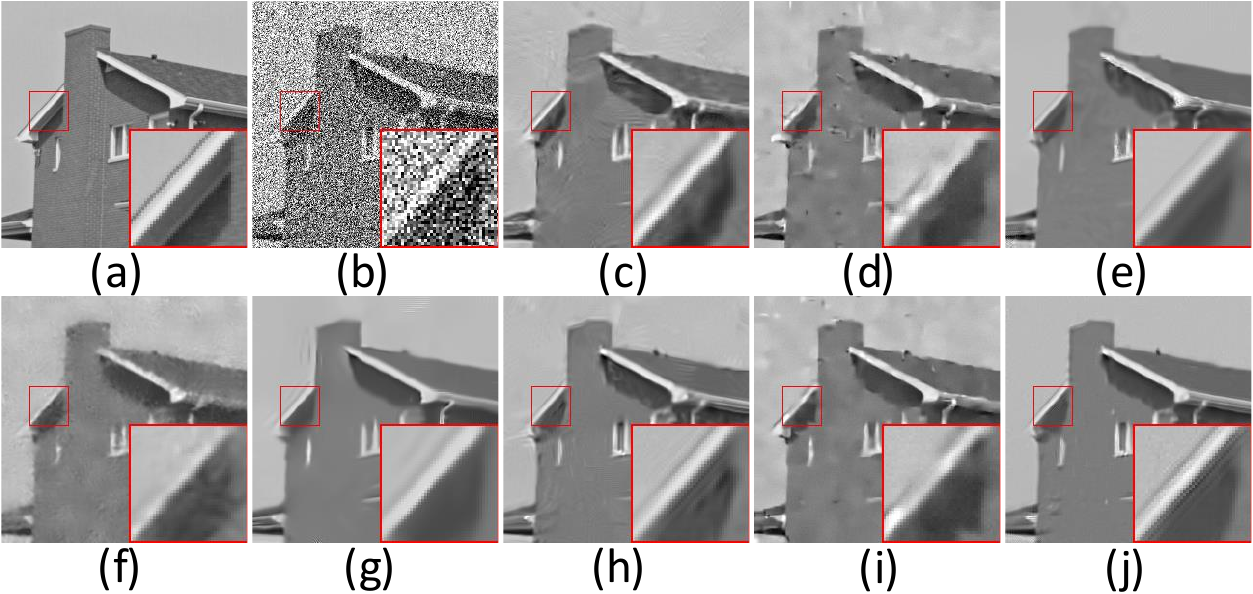}}
\end{minipage}
\caption{Denoising results on image $\emph{House}$ by different methods (noise level $\sigma=75$). (a) Original image; (b) Noisy image;  (c) BM3D \cite{12} (PSNR=27.51dB); (d) EPLL \cite{8} (PSNR=27.09dB); (e) NCSR \cite{7} (PSNR=27.22dB); (f) GID \cite{22} (PSNR=25.23dB); (g) LINE \cite{10} (PSNR=27.22dB); (h) PGPD \cite{11} (PSNR=27.81dB); (i) aGMM \cite{23} (PSNR= 27.31dB); (j) Proposed (PSNR=\textbf{28.24dB}).}
\label{fig:4}
\end{figure}

We present the average PSNR results on six noise levels $\sigma$=20, 30, 40, 50, 75 and 100 in Table~\ref{tab:2}. As can be seen from Table~\ref{tab:2}, the proposed method outperforms the other competing methods. It achieves 0.24dB, 0.59dB, 0.29dB, 1.30dB, 0.28dB, 0.14dB and 0.25dB improvement on average are the BM3D, EPLL, NCSR, GID, LINE, PGPD and aGMM, respectively.
\begin{table}[t]
\caption{Average PSNR results (dB) of the different noise level.}
\centering  
\begin{tabular}{|c|c|c|c|c|c|c|c|}
\hline
\multirow{1}{*}{\textbf{{Noise Level}}}& \textbf{20}  &\textbf{30}  &\textbf{40}  &\textbf{50}   &\textbf{75} &\textbf{100}\\
 \hline
  \multirow{1}{*}{BM3D \cite{12}} & 30.62 & 28.63 & 26.97 & 26.20 & 24.34 & 22.89  \\
\hline
  \multirow{1}{*}{EPLL \cite{8}} & 30.39 & 28.37 & 26.91 & 25.60 & 23.82 & 22.46 \\
\hline
 \multirow{1}{*}{NCSR \cite{7}}  & 30.78 & 28.67 & 27.19 & 26.06 & 24.03 & 22.68 \\
 \hline
 \multirow{1}{*}{GID \cite{22}}   & 29.73 & 27.64 & 26.28 & 25.18 & 23.00 & 21.45  \\
\hline
  \multirow{1}{*}{LINE \cite{10}} & 30.70 & 28.71 & 27.30 & 26.15 & 23.98 & 22.58 \\
\hline
 \multirow{1}{*}{PGPD \cite{11}}  & 30.72 & 28.68 & 27.28 & 26.22 & 24.33 & 23.03 \\
 \hline
 \multirow{1}{*}{aGMM \cite{23}}  & 30.76 & 28.71 & 27.29 & 26.13 & 24.07 & 22.62 \\
 \hline
 \multirow{1}{*}{Proposed} & \textbf{30.81} & \textbf{28.82} & \textbf{27.42} & \textbf{26.34} & \textbf{24.50}& \textbf{23.19}\\
 \hline
\end{tabular}
\label{tab:2}
\end{table}
The visual comparisons of competing denoising methods at noise level 40 and 75 are shown in Fig.~\ref{fig:3} and Fig.~\ref{fig:4}, respectively. It can be seen that BM3D and LINE are resulting in over-smooth phenomena, while EPLL, NCSR, GID, PGPD and aGMM are likely to generate some undesirable ringing artifacts. By contrast, the proposed method is able to preserve the image local structures and suppress undesirable ringing artifacts more effectively than the other competing methods.

\section{Conclusion}
In this paper, we propose a novel method for image denoising using group sparsity residual and external NSS prior. We first propose the concept of the group sparsity residual, and thus the problem of image denoising is turned into reducing the group sparsity residual. To reduce residual, we achieve a good estimation of the group sparse coefficients of the original image by the NSS prior of natural images based on GMM learning and the group sparse coefficients of noisy input image is exploited to approximate this estimation. Experimental results have demonstrated that the proposed method can not only lead to visual improvements over many state-of-the-art methods, but also preserve much better the image local structures and generate much less ringing artifacts.


\begin{thebibliography}{99}
\bibitem{1}
Buades A, Coll B, Morel J M. A non-local algorithm for image denoising[C]//2005 IEEE Computer Society Conference on Computer Vision and Pattern Recognition (CVPR'05). IEEE, 2005, 2: 60-65.
\bibitem{2}
Tikhonov A N, Glasko V B. Use of the regularization method in non-linear problems[J]. USSR Computational Mathematics and Mathematical Physics, 1965, 5(3): 93-107.
\bibitem{3}
Rudin L I, Osher S, Fatemi E. Nonlinear total variation based noise removal algorithms[J]. Physica D: Nonlinear Phenomena, 1992, 60(1): 259-268.
\bibitem{4}
Zhang J, Zhao D, Xiong R, et al. Image restoration using joint statistical modeling in a space-transform domain[J]. IEEE Transactions on Circuits and Systems for Video Technology, 2014, 24(6): 915-928.
\bibitem{5}
Dong W, Zhang L, Shi G, et al. Image deblurring and super-resolution by adaptive sparse domain selection and adaptive regularization[J]. IEEE Transactions on Image Processing, 2011, 20(7): 1838-1857.
\bibitem{6}
Elad M, Aharon M. Image denoising via sparse and redundant representations over learned dictionaries[J]. IEEE Transactions on Image processing, 2006, 15(12): 3736-3745.
\bibitem{7}
Dong W, Zhang L, Shi G, et al. Nonlocally centralized sparse representation for image restoration[J]. IEEE Transactions on Image Processing, 2013, 22(4): 1620-1630.
\bibitem{8}
Zoran D, Weiss Y. From learning models of natural image patches to whole image restoration[C]//2011 International Conference on Computer Vision. IEEE, 2011: 479-486.
\bibitem{9}
Yu G, Sapiro G, Mallat S. Solving inverse problems with piecewise linear estimators: From Gaussian mixture models to structured sparsity[J]. IEEE Transactions on Image Processing, 2012, 21(5): 2481-2499.
\bibitem{10}
Niknejad M, Rabbani H, Babaie-Zadeh M. Image Restoration Using Gaussian Mixture Models With Spatially Constrained Patch Clustering[J]. IEEE Transactions on Image Processing, 2015, 24(11): 3624-3636.
\bibitem{11}
Xu J, Zhang L, Zuo W, et al. Patch group based nonlocal self-similarity prior learning for image denoising[C]//Proceedings of the IEEE International Conference on Computer Vision. 2015: 244-252.
\bibitem{12}
Dabov K, Foi A, Katkovnik V, et al. Image denoising by sparse 3-D transform-domain collaborative filtering[J]. IEEE Transactions on image processing, 2007, 16(8): 2080-2095.
\bibitem{13}
Mairal J, Bach F, Ponce J, et al. Non-local sparse models for image restoration[C]//2009 IEEE 12th International Conference on Computer Vision. IEEE, 2009: 2272-2279.
\bibitem{14}
Zhang L, Dong W, Zhang D, et al. Two-stage image denoising by principal component analysis with local pixel grouping[J]. Pattern Recognition, 2010, 43(4): 1531-1549.
\bibitem{15}
Dong W, Shi G, Ma Y, et al. Image Restoration via Simultaneous Sparse Coding: Where Structured Sparsity Meets Gaussian Scale Mixture[J]. International Journal of Computer Vision, 2015, 114(2-3): 217-232.
\bibitem{16}
Zhang J, Zhao D, Gao W. Group-based sparse representation for image restoration[J]. IEEE Transactions on Image Processing, 2014, 23(8): 3336-3351.
\bibitem{17}
Bishop C M. Pattern recognition[J]. Machine Learning, 2006, 128.
\bibitem{18}
Sandeep P, Jacob T. Single Image Super-Resolution Using a Joint GMM Method[J]. IEEE Transactions on Image Processing, 2016, 25(9): 4233-4244.
\bibitem{19}
Daubechies I, Defrise M, De Mol C. An iterative thresholding algorithm for linear inverse problems with a sparsity constraint[J]. Communications on pure and applied mathematics, 2004, 57(11): 1413-1457.
\bibitem{20}
Chang S G, Yu B, Vetterli M. Adaptive wavelet thresholding for image denoising and compression[J]. IEEE transactions on image processing, 2000, 9(9): 1532-1546.
\bibitem{21}
Osher S, Burger M, Goldfarb D, et al. An iterative regularization method for total variation-based image restoration[J]. Multiscale Modeling
\& Simulation, 2005, 4(2): 460-489.
\bibitem{22}
Talebi H, Milanfar P. Global image denoising[J]. IEEE Transactions on Image Processing, 2014, 23(2): 755-768.
\bibitem{23}
Luo E, Chan S H, Nguyen T Q. Adaptive Image Denoising by Mixture Adaptation[J]. IEEE Transactions on Image Processing, 2016, 25(10): 4489-4503.
\bibitem{24}
Dong W, Shi G, Li X. Nonlocal image restoration with bilateral variance estimation: a low-rank approach[J]. IEEE transactions on image processing, 2013, 22(2): 700-711.
\end{thebibliography}

\end{document}